\documentclass[11pt]{article}

\usepackage[]{acl}

\usepackage{times}
\usepackage{latexsym}
\usepackage{epigraph}
\usepackage{xspace}

\usepackage[T1]{fontenc}

\usepackage[utf8]{inputenc}

\usepackage{microtype}
\usepackage{graphicx}
\usepackage{enumitem} 
\usepackage{caption}
\usepackage{subcaption}
\usepackage{booktabs}
\usepackage{xcolor}
\usepackage{fancybox}
\usepackage[most]{tcolorbox}
\definecolor{ShadowColor}{RGB}{30,150,190}
\colorlet{ShadowColor}{gray}

\newcommand{\method}{\textsc{$\mathcal{R}3$}\xspace}
\usepackage[normalem]{ulem}
\usepackage[finalnew]{trackchanges}%
\addeditor{VR}
\addeditor{WD}
\addeditor{DK}
\addeditor{ZM}
\addeditor{DH}

\title{Read, Revise, Repeat: A System Demonstration for Human-in-the-loop Iterative Text Revision}

\makeatletter
\newcommand{\printfnsymbol}[1]{%
  \textsuperscript{\@fnsymbol{#1}}%
}
\makeatother

\author{Wanyu Du$^1$\thanks{\xspace\xspace Equal contributions.}\ , Zae Myung Kim$^2$\printfnsymbol{1}\ , Vipul Raheja$^3$\ ,  Dhruv Kumar$^3$\ , Dongyeop Kang$^2$ \\
$^1$University of Virginia, \space $^2$University of Minnesota, \space $^3$Grammarly \\
\texttt{wd5jq@virginia.edu}, \space
\texttt{\{kim01756,dongyeop\}@umn.edu} \\
\texttt{\{vipul.raheja,dhruv.kumar\}@grammarly.com}
}

\begin{document}
\maketitle
\begin{abstract}
Revision is an essential part of the human writing process. It tends to be strategic, adaptive, and, more importantly, \textit{iterative} in nature. 
Despite the success of large language models on text revision tasks, they are limited to non-iterative, one-shot revisions. 
Examining and evaluating the capability of large language models for making continuous revisions and collaborating with human writers is a critical step towards building effective writing assistants.
In this work, we present a human-in-the-loop iterative text revision system, 
$\mathcal{R}$ead, $\mathcal{R}$evise, $\mathcal{R}$epeat (\textsc{$\mathcal{R}3$}), which aims at achieving high quality text revisions with minimal human efforts by reading model-generated revisions and user feedbacks, revising documents, and repeating human-machine interactions.
In \method, a text revision model provides text editing suggestions for human writers, who can accept or reject the suggested edits. The accepted edits are then incorporated into the model for the next iteration of document revision.
Writers can therefore revise documents iteratively by interacting with the system and simply accepting/rejecting its suggested edits until the text revision model stops making further revisions or reaches a predefined maximum number of revisions.
Empirical experiments show that \method can generate revisions with comparable acceptance rate to human writers at early revision depths, and the human-machine interaction can get higher quality revisions with fewer iterations and edits. 
The collected human-model interaction dataset and system code are available at \url{https://github.com/vipulraheja/IteraTeR}. 
Our system demonstration is available at \url{https://youtu.be/lK08tIpEoaE}. 
\end{abstract}

\section{Introduction}
\begin{figure}[t]
    \centering
    \includegraphics[width=0.42\textwidth]{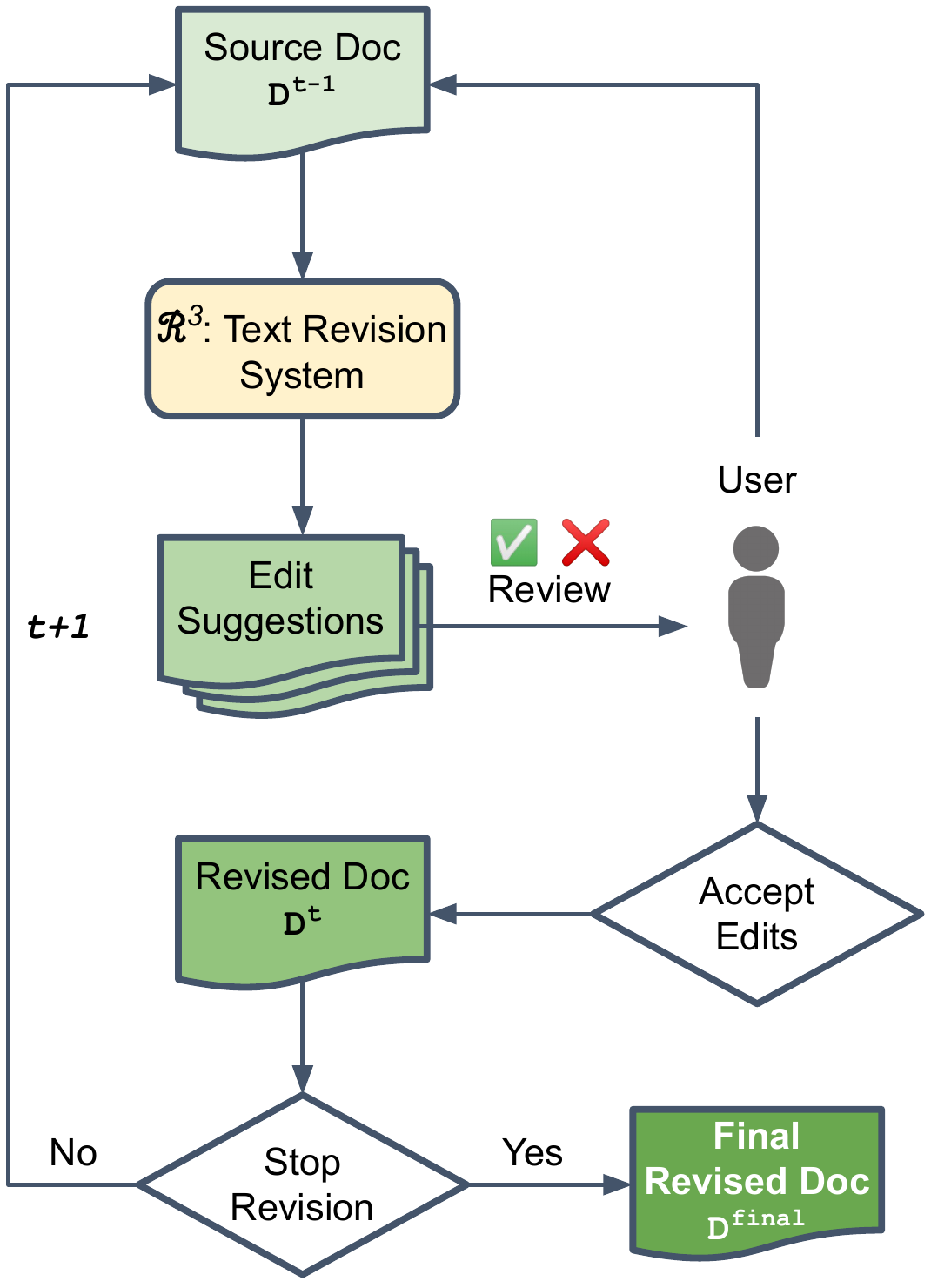}
    \caption{System overview for \method human-in-the-loop iterative text revision.}
    \label{fig:system_overview}
\end{figure}

Text revision is a crucial part of writing. Specifically, text revision involves identifying discrepancies between intended and instantiated text, deciding what edits to make, and how to make those desired edits \cite{10.2307/356600, faigley1981analyzing, 10.2307/1170433}. It enables writers to deliberate over and organize their thoughts, find a better line of argument, learn afresh, and discover what was not known before \cite{sommers1980revision, scardamalia1986}. 
Previous studies \citep{flower1980dynamics, collins1980framework, vaughan-mcdonald-1986-model} have shown that text revision is an \textit{iterative} process since human writers are unable to simultaneously comprehend multiple demands and constraints of the task when producing well-written texts -- for instance, covering the content, following linguistic norms and discourse conventions of written prose, etc. Therefore, writers resort to performing text revisions on their drafts iteratively to reduce the number of considerations at each time.

Computational modeling of the iterative text revision process is essential for building intelligent and interactive writing assistants.
Most prior works on the development of neural text revision systems \citep{faruqui-etal-2018-wikiatomicedits,botha-etal-2018-learning,ito-etal-2019-diamonds,faltings-etal-2021-text} 
do not take the iterative nature of text revision and human feedback on suggested revisions into consideration. 
The direct application of such revision systems in an iterative way, however, could generate some ``noisy'' edits and require much burden on human writers to fix the noise.
Therefore, we propose to collect human feedback at each iteration of revision to filter out those harmful noisy edits and produce revised documents of higher quality.


In this work, we present a novel human-in-the-loop iterative text revision system, $\mathcal{R}$ead, $\mathcal{R}$evise, $\mathcal{R}$epeat (\method), which reads model-generated revisions and user feedbacks, revises documents, and repeats human-machine interactions in an iterative way, as depicted in \autoref{fig:system_overview}.
First, users write a document as input to the system or choose one from a candidate document set to edit.
Then, the text revision system provides multiple editing suggestions with their edits and intents.
Users can accept or reject the editing suggestions in an iterative way and stop revision when no editing suggestions are provided or the model reaches the maximum revision limit.
The overall model performance can be estimated by calculating the acceptance rate throughout all editing suggestions.

\method provides numerous benefits over existing writing assistants for text revision. 
First, \method improves the overall writing experience for writers by making it more interpretable, controllable, and productive: on the one hand, writers don't have to (re-)read the parts of the text that are already high quality, and this, in turn, helps them focus on larger writing goals (\S \ref{sec:results}); on the other hand, by showing edit intentions for every suggested edit, which users can further decide to accept or reject, \method provides them with more fine-grained control over the text revision process compared to other one-shot based text revision systems \cite{lee2022coauthor}, and are limited in both interpretability and controllability. 
Second, \method improves the revision efficiency.
The human-machine interaction can help the system produce higher quality revisions with fewer iterations and edits, and the empirical experiments in \S \ref{sec:results} validate this claim.
To the best of our knowledge, \method is the first text revision system in literature that can perform \textit{iterative} text revision in collaboration by human writers and revision models.

In this paper, we make three major contributions:
\begin{itemize}
    \item We present a novel human-in-the-loop text revision system \method to make text revision models more accessible; and to make the process of iterative text revision efficient, productive, and cognitively less challenging.
    \item From an HCI perspective, we conduct experiments to measure the effectiveness of the proposed system for the iterative text revision task. Empirical experiments show that \method can generate edits with comparable acceptance rate to human writers at early revision depths. 
    \item We analyze the data collected from human-model interactions for text revision and provide insights and future directions for building high-quality and efficient human-in-the-loop text revision systems. We release our code, revision interface, and collected human-model interaction dataset to promote future research on collaborative text revision.
\end{itemize}

\section{Related Work}
Previous works on modeling text revision \citep{faruqui-etal-2018-wikiatomicedits,botha-etal-2018-learning,ito-etal-2019-diamonds,faltings-etal-2021-text} have ignored the iterative nature of the task, and simplified it into a one-shot "original-to-final" sentence-to-sentence generation task. However, in practice, at every revision step, multiple edits happen at the document-level which also play an important role in text revision. For instance, reordering and deleting sentences to improve the coherence. 

More importantly, performing multiple high-quality edits at once is very challenging. 
Continuing the previous example, document readability can degrade after reordering sentences, and further adding transitional phrases is often required to make the document more coherent and readable. 
Therefore, one-shot sentence-to-sentence text revision formulation is not sufficient to deal with real-world challenges in text revision tasks.

While some prior works on text revision \citep{coenen2021wordcraft,padmakumar2021machine,gero2021sparks,lee2022coauthor} have proposed human-machine collaborative writing interfaces, they are mostly focused on collecting human-machine interaction data for training better neural models, rather than understanding the iterative nature of the text revision process, or the model's ability to adjust editing suggestions according to human feedback.

Another line of work by \citet{sun-etal-2021-iga,10.1145/3511599} on creative writing designed human-machine interaction interfaces to encourage new content generation.
However, text revision focuses on improving the quality of existing writing and keeping the original content as much as possible.
In this work, we provide a human-in-the-loop text revision system to make helpful editing suggestions by interacting with users in an iterative way.

\section{System Overview}
\autoref{fig:system_overview} shows the general pipeline of \method human-in-the-loop iterative text revision system.
In this section, we will describe the development details of the text revision models and demonstrate our user interfaces.

\begin{figure*}[t]
    \centering
     \begin{subfigure}[b]{0.22\textwidth}
         \centering
         \begin{tcolorbox}[enhanced jigsaw,drop shadow=black!50!white,colback=white,size=tight,boxrule=1pt]
         \includegraphics[width=\textwidth,height=2.8cm]{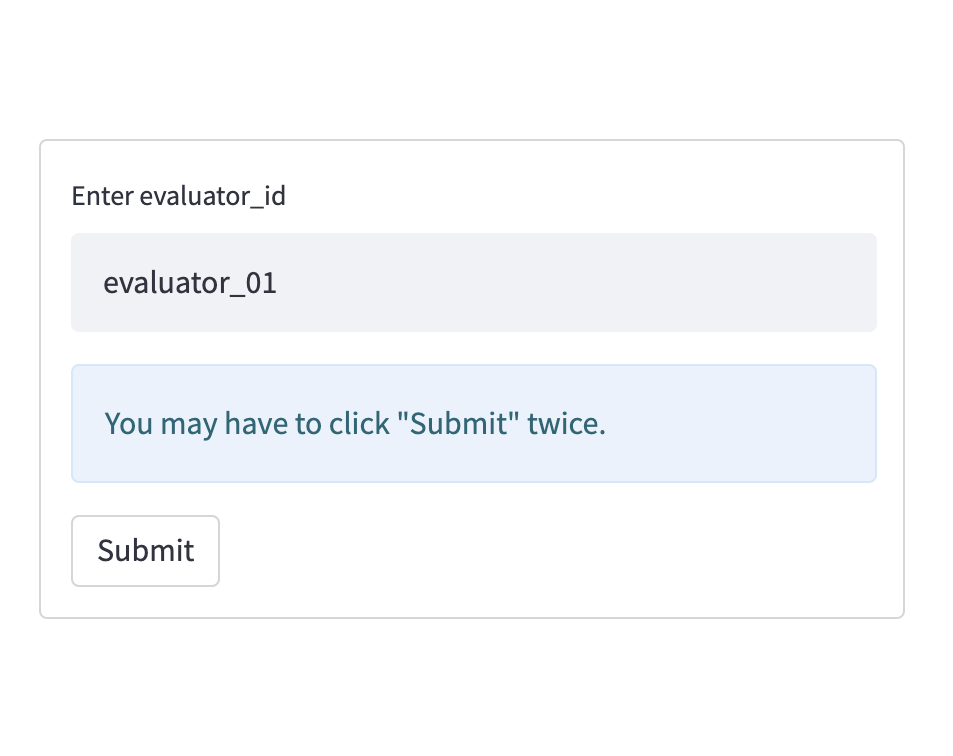}
         \end{tcolorbox}
         \caption{Login}
         \label{fig:login}
     \end{subfigure}
     \hfill
     \begin{subfigure}[b]{0.53\textwidth}
         \centering
         \begin{tcolorbox}[enhanced jigsaw,drop shadow=black!50!white,colback=white,size=tight,boxrule=1pt]
         \includegraphics[width=\textwidth,height=3.6cm]{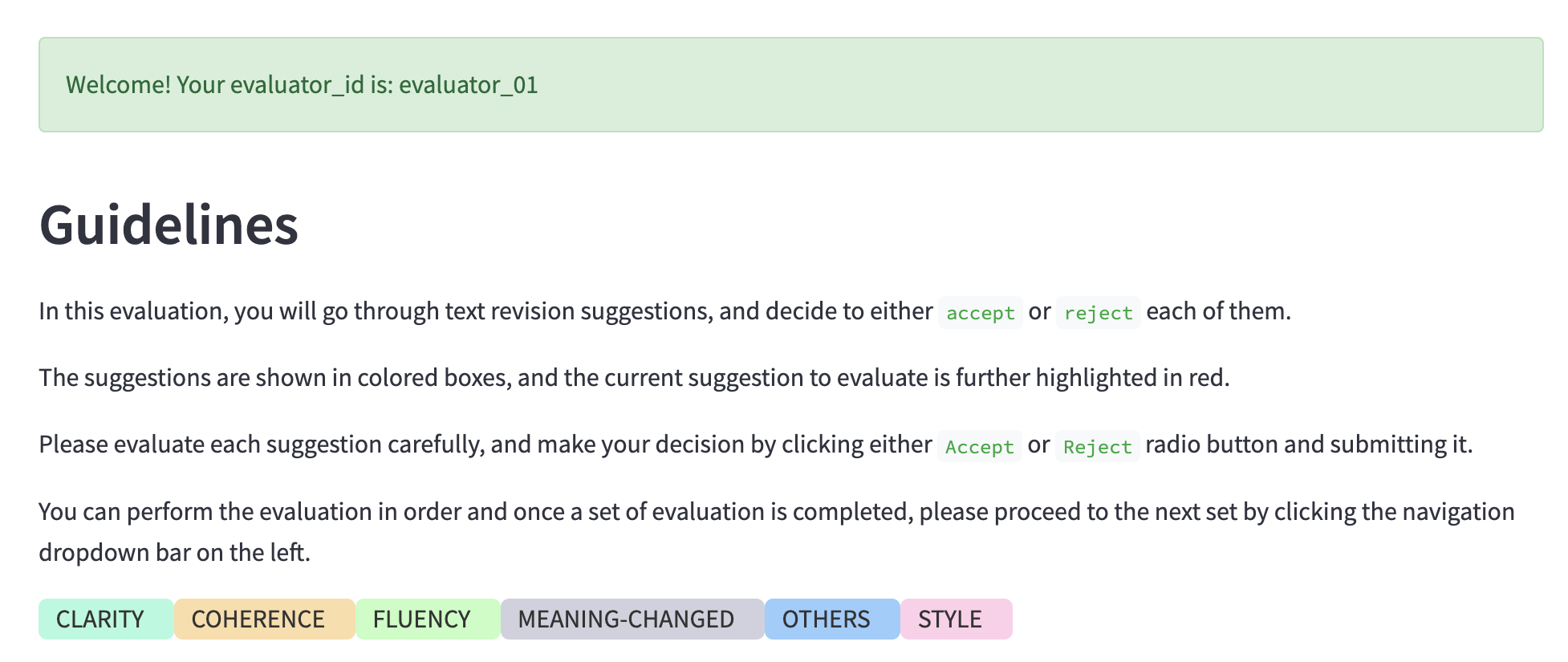}
         \end{tcolorbox}
         \caption{Read guidelines}
         \label{fig:guideline}
     \end{subfigure}
     \hfill
     \begin{subfigure}[b]{0.17\textwidth}
         \centering
         \begin{tcolorbox}[enhanced jigsaw,drop shadow=black!50!white,colback=white,size=tight,boxrule=1pt]
         \includegraphics[width=\textwidth,height=3.6cm]{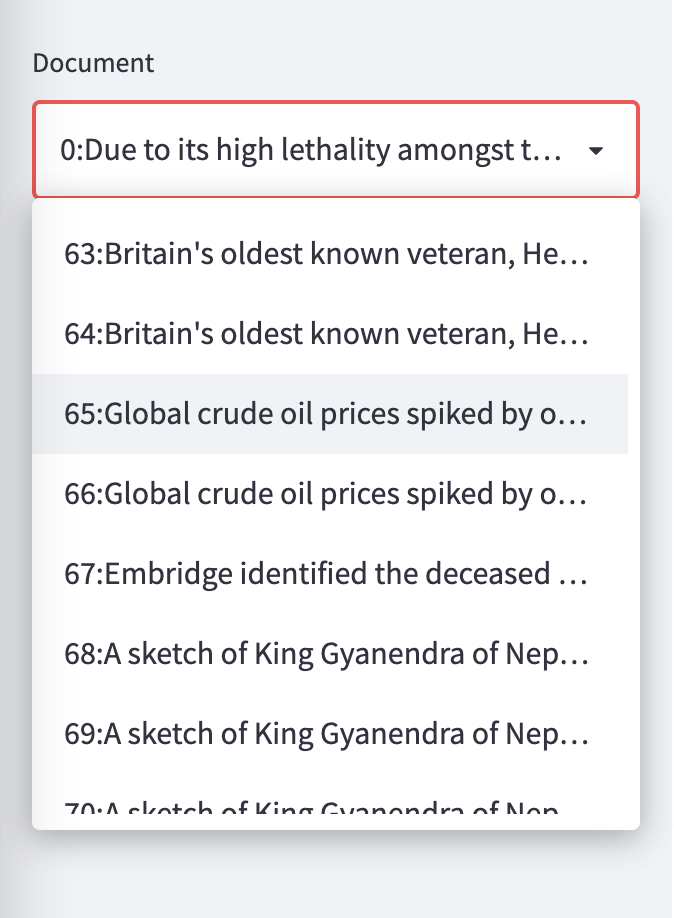}
         \end{tcolorbox}
         \caption{Select doc}
         \label{fig:select_doc}
     \end{subfigure}
     \newline
     \newline
     \begin{subfigure}[b]{0.99\textwidth}
         \centering
         \begin{tcolorbox}[enhanced jigsaw,drop shadow=black!50!white,colback=white]
         \includegraphics[width=\textwidth,trim={0 0 0 0},clip]{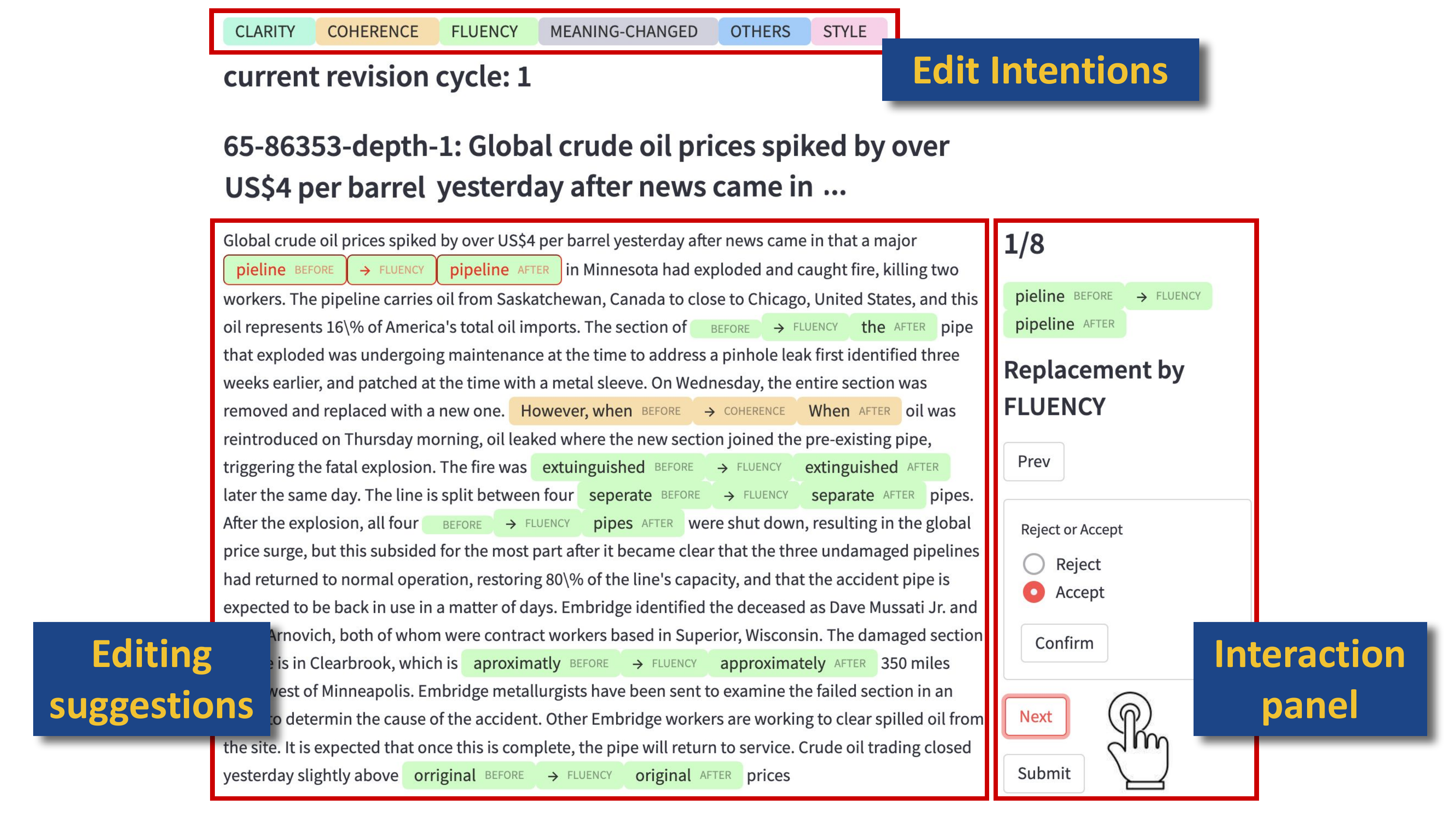}
         \end{tcolorbox}
         \caption{Editing suggestions and interaction panel}
         \label{fig:edit}
     \end{subfigure}     
    \caption{User interface demonstration for \method. Anonymized version available at \url{https://youtu.be/lK08tIpEoaE}.}
    \label{fig:user_demo}
\end{figure*}

We first formulate an iterative text revision process:
given a source document\footnote{The source document can be chosen by a user in the candidate set of documents or written from scratch by a user.} $\mathcal{D}^{t-1}$, at each revision depth $t$, a text revision system will apply a set of edits to get the revised document $\mathcal{D}^{t}$.
The system will continue iterating revision until the revised document $\mathcal{D}^{t}$ satisfies a set of predefined stopping criteria, such as reaching a predefined maximum revision depth $t_{max}$, or making no edits between $\mathcal{D}^{t-1}$ and $\mathcal{D}^{t}$.

\subsection{Text Revision System}

We follow the prior work of \citet{du-etal-2022-understanding-iterative} to build our text revision system.
The system is composed of edit intention identification models and a text revision generation model.
We follow the same data collection procedure in \citet{du-etal-2022-understanding-iterative} to collect the iterative revision data.\footnote{See \S \ref{sec:setup} for the detailed data collection.}
Then, we train the three models on the collected revision dataset.

\paragraph{Edit Intention Identification Models.}
Following \citet{du-etal-2022-understanding-iterative}, our edit intentions have four categories: \textsc{fluency}, \textsc{coherence}, \textsc{clarity}, and \textsc{style}.
We build our edit intention identification models at each sentence of the source document $\mathcal{D}^{t-1}$ to capture the more fine-grained edits.
Specifically, given a source sentence, the system will make two-step predictions: (1) whether or not to edit, and (2) which edit intention to apply. The decision whether or not to edit is taken by an edit-prediction classifier that predicts a binary label of whether to edit a sentence or not. The second model, called the edit-intention classifier, predicts which edit intention to apply to the sentence.
If the edit-prediction model predicts ``not to edit'' in the first step, the source sentence will be kept unchanged at the current revision depth. 


\paragraph{Text Revision Generation Model.}
We fine-tune a large pre-trained language model like \textsc{Pegasus} \citep{pmlr-v119-zhang20ae} on our collected revision dataset to build the text revision generation model. 
Given a source sentence and its predicted edit intention, the model will generate a revised sentence, conditioned on the predicted edit intention. 
Then, we concatenate all un-revised and revised sentences to get the model-revised document $\mathcal{D}^{t}$, and extract all its edits using \textit{latexdiff}\footnote{\url{https://ctan.org/pkg/latexdiff}} and \textit{difflib}.\footnote{\url{https://docs.python.org/3/library/difflib.html}}

In summary, at each revision depth $t$, given a source document $\mathcal{D}^{t-1}$, the text revision system first predicts the need for revising a sentence, and for the ones that need revision, it predicts the corresponding fine-grained edit intentions -- thus, generating the revised document $\mathcal{D}^{t}$ based on the source document and the predicted edit decisions and intentions.

\subsection{Human-in-the-loop Revision}
In practice, not all model-generated edits are equally impactful towards improving the document quality \cite{du-etal-2022-understanding-iterative}. Therefore, we enable user interaction in the iterative text revision process to achieve high quality of text revisions along with a productive writing experience.
At each revision depth $t$, our system provides the user with suggested edits, and their corresponding edit intentions.  
The user can interact with the system by choosing to accept or reject the suggested edits. 

\autoref{fig:user_demo} illustrates the details of \method's user interface.
First, a user enters their id to login to the web interface as shown in \autoref{fig:login}.
Then, the user is instructed with a few guidelines on how to operate the revision as demonstrated in \autoref{fig:guideline}.
After getting familiar with the interface, the user can select a source document from the left drop-down menu in \autoref{fig:select_doc}.
By clicking the source document, all the edits predicted by the text revision model, as well as their corresponding edit intentions will show up in the main page as illustrated in \autoref{fig:edit} (left panel).
The user is guided to go through each suggested edits, and choose to accept or reject the current edit by clicking the \textit{Confirm} button in \autoref{fig:edit} (right panel).
After going through all the suggested edits, the user is guided to click the \textit{Submit} button to save their decisions on the edits.
Then, the user is guided to click the \textit{Next Iteration!} button to proceed to the next revision depth and check the next round of edits suggested by the system.
This interactive process continues until the system does not generate further edits or reaches the maximum revision depth $t_{max}$.

\section{Experiments}
We conduct experiments to answer the following research questions: 
\begin{enumerate}[noitemsep,topsep=0pt,leftmargin=*,label=RQ\arabic*]
    \item How likely are users to accept the editing suggestions predicted by our text revision system? 
    \add[WD]{This question is designed to evaluate whether our text revision system can generate high quality edits. }
    \item Which types of edit intentions are more likely to be accepted by users?
    \add[WD]{This question is aimed to identify which types of edits are more favored by users.}
    \item Does user feedback in \method help produce higher quality of revised documents?
    \add[WD]{This question is proposed to validate the effectiveness of human-in-the-loop component in \method. }
\end{enumerate}

\subsection{Experimental Setups}\label{sec:setup}

\paragraph{Iterative Revision Systems.}
We prepare three types of iterative revision systems to answer the above questions:
\begin{enumerate}[noitemsep,topsep=0pt,leftmargin=*]
    \item \textsc{Human-Human}: We ask users to accept or reject text revisions made by human writers, which are directly sampled from our collected iterative revision dataset. This serves as the baseline to measure the gap between our text revision system and human writers.
    \item \textsc{System-Human}: We ask users to accept or reject text revisions made by our system. Then, we incorporate user accepted edits to the system to generate the next iteration of revision. This is the standard human-in-the-loop process of \method.
    \item \textsc{System-Only}: We conduct an ablation study by removing user interaction in reviewing the model-generated edits. Then, we compare the overall quality of final revised documents with and without the human-in-the-loop component.   
\end{enumerate}
In both \textsc{Human-Human} and \textsc{System-Human} setups where users interacted with the system, they were not informed whether the revisions were sampled from our collected iterative revision dataset, or generated by the underlying text revision models.

\paragraph{User Study Design.}
We hired three linguistic experts (English L1, bachelor's or higher degree in Linguistics) to interact with our text revision system.
Each user was presented with a text revision (as shown in \autoref{fig:edit}) and asked to accept or reject each edit in the current revision \add[WD]{(users were informed which revision depth they were looking at)}.
For a fair comparison, users were not informed about the source of the edits (human-written vs. model-generated), and the experiments were conducted separately one after the other.
Note that the users were only asked to accept or reject edits, and they had control neither over the number of iterations, nor over the stopping criteria.
The stopping criteria for the experiment were \add[ZM]{set by us and }designed as: (1) no new edits were made at the following revision depth, or (2) the maximum revision depth $t_{max}=3$ was reached.

\begin{table}[t]
  \centering
  \small
  \begin{tabular}{@{}lrcr@{}}
    \toprule
    & \textbf{\# Docs} & \textbf{Avg. Depths} & \textbf{\# Edits} \\
    \midrule
    Training & 44,270 & 6.63 & 292,929 \\
    Validation & 5,152 & 6.60 & 34,026 \\
    Test & 6,226 & 6.34 & 39,511 \\
    \bottomrule
  \end{tabular}
  \caption{\label{tab:data}
  Statistics for our collected revision data which has been used to train the edit intention identification model and the text revision generation model. \textbf{\# Docs} means the total number of unique documents, \textbf{Avg. Depths} indicates the average revision depth per document\add[ZM]{ (for the human-generated training data)}, and \textbf{\# Edits} stands for the total number of edits (sentence pairs) across the corpus.
  }
\end{table}
\begin{table*}[th]
  \centering
  \small
  \begin{tabular}{@{}lcccccccc@{}}
    \toprule
    & \multicolumn{4}{c}{\textsc{Human-Human}} & \multicolumn{4}{c}{\textsc{System-Human} (ours)} \\
    \cmidrule(lr){2-5} \cmidrule(lr){6-9}
    \textbf{$t$} & \textbf{\# Docs} & \textbf{Avg. Edits} & \textbf{Avg. Accepts} & \textbf{\% Accepts} & \textbf{\# Docs} & \textbf{Avg. Edits} & \textbf{Avg. Accepts} & \textbf{\% Accepts} \\
    \midrule
    1 & 30 & 5.37 & 2.77 & 51.66 & 30 & \textbf{5.90} & \textbf{2.90} & 49.15 \\
    2 & 30 & 4.83 & 3.00 & 62.06 & 24 & 3.83 & 2.57 & \textbf{67.02} \\
    3 & 20 & 3.80 & 2.67 & 70.39 & 20 & 3.43 & 1.94 & 56.71 \\
    \bottomrule
  \end{tabular}
  \caption{\label{tab:eval}
  Human-in-the-loop iterative text revision evaluation results. \textbf{$t$} stands for the revision depth, \textbf{\# Docs} shows the total number of revised documents at the current revision depth, \textbf{Avg. Edits} indicates the average number of applied edits per document, \textbf{Avg. Accepts} means the average number of edits accepted by users per document, and \textbf{\% Accepts} is calculated by dividing the total accepted edits with the total applied edits.
  }
\end{table*}

\paragraph{Data Details.}
We followed the prior work \citep{du-etal-2022-understanding-iterative} to collect the text revision data across three domains: ArXiv, Wikipedia and Wikinews.
This data was then used to train both the edit intention identification models and the text revision generation model.
We split the data into training, validation and test set according to their document ids with a ratio of 8:1:1.
The detailed data statistics are included in \autoref{tab:data}.
Note that our newly collected revision dataset is larger than the previously proposed dataset in \citet{du-etal-2022-understanding-iterative} with around 24K more unique documents and 170K more edits (sentence pairs).

For the human evaluation data, we randomly sampled 10 documents with a maximum revision depth of 3 from each domain in the test set in \autoref{tab:data}.
For the evaluation of text revisions made by human writers (\textsc{Human-Human}), we presented the existing ground-truth references from our collected dataset to users.
Since we do not hire additional human writers to perform continuous revisions, we just presented the static human revisions from the original test set to users at each revision depth, and collected the user acceptance statistics as a baseline for our system. 

For the evaluation of text revisions made by our system (\textsc{System-Human}), we only presented the original source document at the initial revision depth ($\mathcal{D}^{0}$) to our system, and let the system generate edits in the following revision depths, while incorporating the accept/reject decisions on model-generated edit suggestions by the users. 
Note that at each revision depth, the system will only incorporate the edits accepted by users and pass them to the next revision iteration.

For text revisions made by our system without human-in-the-loop (\textsc{System-Only}), we let the system generate edits in an iterative way and accepted all model-generated edits at each revision depth.

\paragraph{Model Details.}
For both edit intention identification models, we fine-tuned the RoBERTa-large \citep{liu2020roberta} pre-trained checkpoint from HuggingFace \citep{wolf-etal-2020-transformers} for 2 epochs with a learning rate of $1\times 10^{-5}$ and batch size of 16. The edit-prediction classifier is binary classification model that predicts whether to edit a given sentence or not. It achieves an F1 score of 67.33 for the edit label and 79.67 for the not-edit label. The edit-intention classifier predicts the specific intent for a sentence that requires editing. It achieves F1 scores of 67.14, 70.27, 57.0, and 3.21\add[ZM]{\footnote{We note that the F1 score for \textsc{style} is low as the number of training samples for that intent is particularly small.}} for \textsc{clarity}, \textsc{fluency}, \textsc{coherence} and \textsc{style} intent labels respectively.

For the text revision generation model, we fine-tuned the \textsc{Pegasus-large} \citep{pmlr-v119-zhang20ae} pre-trained checkpoint from HuggingFace.
We set the edit intentions as new special tokens (e.g., \texttt{<STYLE>}, \texttt{<FLUENCY>}), and concatenated the edit intention and source sentence together as the input to the model.
The output of the model is the revised sentence, and we trained the model with cross-entropy loss.
We fine-tuned the model for 5 epochs with a learning rate of $3\times 10^{-5}$ and batch size of 4.  
Finally, our text revision generation model achieves 41.78 SARI score \citep{xu-etal-2016-optimizing}, 81.11 BLEU score \citep{papineni-etal-2002-bleu} and 89.08 ROUGE-L score \citep{lin-2004-rouge} on the test set.

\begin{table*}[th]
  \centering
  \small
  \begin{tabular}{@{}lrrrrrr@{}}
    \toprule
    & \multicolumn{3}{c}{\textsc{Human-Human}} & \multicolumn{3}{c}{\textsc{System-Human} (ours)} \\
    \cmidrule(lr){2-4} \cmidrule(lr){5-7}
    & \textbf{\# Edits} & \textbf{\# Accepts} & \textbf{\% Accepts} & \textbf{\# Edits} & \textbf{\# Accepts} & \textbf{\% Accepts} \\
    \midrule
    \textsc{Clarity} & 197 & 119 & 60.40 & \textbf{332} & 195 & 58.73 \\
    \textsc{Fluency} & 178 & 146 & \textbf{82.02} & 91 & 41 & 45.05 \\
    \textsc{Coherence} & 103 & 41 & 39.80 & 141 & 68 & 48.22 \\
    \textsc{Style} & 6 & 2 & 33.33 & 113 & 73 & \textbf{64.60} \\
    \bottomrule
  \end{tabular}
  \caption{\label{tab:eval_intents}
  The distribution of different edit intentions.
  \textbf{\# Edits} indicates the total number of applied edits under the current edit intention, \textbf{\# Accepts} means the total number of edits accepted by users under the current edit intention, and \textbf{\% Accepts} is calculated by dividing the total accepted edits with the total applied edits.
  }
\end{table*}
\begin{table}[th]
  \centering
  \small
  \begin{tabular}{lp{0.31\textwidth}}
    \toprule
     \textbf{Edit Intention} & \textbf{Edit Suggestion}  \\
    \toprule
    \textsc{Clarity} & Emerging new test procedures \textcolor{red}{\sout{, such as antigen or RT-LAMP tests,}} might enable us to protect nursing home residents. \\
    \midrule
    \textsc{Fluency} & For Radar tracking\textcolor{teal}{,} we show how a model can reduce the tracking errors. \\
    \midrule
    \textsc{Coherence} & However, \textcolor{red}{\sout{we show that}} even a small violation can significantly modify the effective noise.\\
    \midrule
    \textsc{Style} & There has been \textcolor{red}{\sout{numerous}}\textcolor{teal}{extensive} research focusing on neural coding.\\
    \bottomrule
  \end{tabular}
  \caption{\label{tab:edit_examples}
  Edit suggestion examples generated by \method. 
  }
\end{table}

\begin{table}[t]
  \centering
  \small
  \begin{tabular}{@{}lc@{\hskip 2mm}c@{\hskip 2mm}c@{}}
    \toprule
    & \textbf{Avg. Depths} & \textbf{\# Edits} & \textbf{Quality} \\
    \midrule
    \textsc{System-Human} (ours) & 2.5 & 148 & \textbf{0.68} \\
    \textsc{System-Only} & 2.8 & 175 & 0.28 \\
    \bottomrule
  \end{tabular}
  \caption{\label{tab:quality}
  Quality comparison results of final revised documents with and without human-in-the-loop.
  \textbf{Avg. Depths} indicates the average number of iterations conducted by the system, \textbf{\# Edits} means the total number of accepted edits by the system, and \textbf{Quality} represents the human judgements of the overall quality of system-revised final documents.
  }
\end{table}
\begin{table}[t]
  \centering
  \small
  \begin{tabular}{@{}lcc@{}}
    \toprule
    \textbf{Criterion} & \textbf{Avg. Score} & \textbf{Std. Deviation}\\
    \midrule
    Convenience & 3.66  & 0.58 \\
    Satisfaction & 2.33  & 0.58 \\
    Productivity & 3.00  & 1.00 \\
    Retention & 2.66  & 0.58 \\
    \bottomrule
  \end{tabular}
  \caption{\label{tab:ux}
  User feedback survey ratings. Ratings are on 5-point Likert scale with 5 being strongly positive experience, 3 being neutral, and 1 being strongly negative.\add[ZM]{ However, we'd like to point out that as the number of users (linguists) who participated in the study is small, the statistical significance of the results should be taken lightly.}}
\end{table}

\subsection{Result Analysis}\label{sec:results}
\paragraph{Iterativeness.}
The human-in-the-loop iterative text revision evaluation results are reported in \autoref{tab:eval}.
Each document is evaluated by at least 2 users.
\textbf{We find that \method achieves comparable performances with ground-truth human revisions at revision depth 1 and 2, and tends to generate less favorable edits at revision depth 3.}
At revision depth 1, \method is able to generate more edits than ground-truth human edits for each document, and gets more edits accepted by users on average.
This shows the potential of \method in generating appropriate text revisions that are more favorable to users.

At revision depth 2, while \method generates less edits than human writers on average, it gets a higher acceptance rate than human writers.
This result suggests that for the end users, more edits may not necessarily lead to a higher acceptance ratio, and shows that \method is able to make high-quality edits for effective iterative text revisions.
At revision depth 3, \method generates even less edits compared both to human writers and its previous revision depths.
This result can be attributed to the fact that our models are only trained on static human revision data, while at testing time they have to make predictions conditioned on their revisions generated at the previous depth, which may have a very different distribution of edits than the training data.
\autoref{tab:revision_example_arxiv} shows an example of iterative text revision in ArXiv domain generated by \method.
\add[WD]{We also provide some other iterative revision examples generated by} \method in \autoref{sec:appendix}.

\paragraph{Edit Intentions.}
\autoref{tab:eval_intents} demonstrates the distribution of different edit intentions, which can help us further analyze the which type of edits are more likely to be accepted by end users.
For human-generated revisions, we find that \textsc{Fluency} edits are most likely to be accepted since they are mainly fixing grammatical errors.

\textbf{For system-generated revisions, we observe that \textsc{Clarity} edits are the most frequent edits but end users only accept 58.73\% of them}, which suggests that our system needs further improvements in learning \textsc{Clarity} edits.
Another interesting observation is that \textsc{Style} edits are rarely generated by human writers \add[ZM]{(1.2\%) }and also gets the lowest acceptance rate \add[ZM]{(33.33\%) }than other intentions, while they are frequently generated by our system \add[ZM]{(16.7\%)} and surprisingly gets the highest acceptance rate \add[ZM]{(64.6\%)} than other intentions. 
This observation indicates that \method is capable for generating favorable stylistic edits. 
\autoref{tab:edit_examples} \add[WD]{shows some examples of edit suggestions generated by \method.}

\paragraph{Role of Human Feedback in Revision Quality.}
\autoref{tab:quality} illustrates the quality comparison results of final revised documents with and without human-in-the-loop for \method.
We asked another group of three annotators (English L2, bachelor's or higher degree in Computer Science) to judge whether the overall quality of system-generated final document is better than the ground-truth reference final document. 
The quality score ranges between 0 and 1.
We evaluated 10 unique documents in ArXiv domain, and took the average score from all 3 annotators.
As shown in \autoref{tab:quality}, \textbf{\textsc{System-Human} produces better overall quality score for the final system-generated documents with fewer iterations of revision and fewer edits}, which validates the effectiveness of the human-machine interaction proposed in \method. 

\begin{table*}[h!]
  \centering
  \small
  \begin{tabular}{@{}r|p{0.4\textwidth}|p{0.4\textwidth}@{}}
    \toprule
     $t$ & \textsc{Human-Human} & \textsc{System-Human} (ours)  \\
    \midrule
    0 
    & Due to its high lethality amongst the elderly, nursing homes are in the eye of the COVID-19 storm. Emerging new test procedures , such as antigen or RT-LAMP tests, might enable us to protect nursing home residents by means of preventive screening strategies. Here, we develop a novel agent-based epidemiological model for the spread of SARS-CoV-2 in nursing homes to identify optimal preventive testing strategiesto curb this spread . The model is microscopically calibrated to high-resolution data from actual nursing homes in Austria, including the detailed networks of social contacts of their residents and information on past outbreaks.
    & Due to its high lethality amongst the elderly, nursing homes are in the eye of the COVID-19 storm. Emerging new test procedures , such as antigen or RT-LAMP tests, might enable us to protect nursing home residents by means of preventive screening strategies. Here, we develop a novel agent-based epidemiological model for the spread of SARS-CoV-2 in nursing homes to identify optimal preventive testing strategiesto curb this spread . The model is microscopically calibrated to high-resolution data from actual nursing homes in Austria, including the detailed networks of social contacts of their residents and information on past outbreaks. \\
    \midrule
    1 
    & Due to its high lethality amongst the elderly, nursing homes are in the eye of the COVID-19 storm. \colorbox{gray!20}{\sout{Emerging new}}\colorbox{gray!20}{With} test procedures \colorbox{gray!20}{becoming available at scale }, such as antigen or RT-LAMP tests, might enable us to protect nursing home residents by means of preventive screening strategies. Here, we develop a novel agent-based epidemiological model for the spread of SARS-CoV-2 in nursing homes to identify optimal \colorbox{gray!20}{\sout{preventive testing strategiesto curb this spread }}
    \colorbox{gray!20}{prevention strategies}. The model is microscopically calibrated to high-resolution data from actual nursing homes in Austria, including the detailed networks of social contacts of their residents and information on past outbreaks. 
    & Due to its high lethality amongst the elderly, nursing homes are in the eye of the COVID-19 storm. Emerging new test procedures\colorbox{red!50}{\sout{ , such as antigen or RT-LAMP tests,}} might enable us to protect nursing home residents by means of preventive screening strategies. Here, we develop a novel agent-based epidemiological model for the spread of SARS-CoV-2 in nursing homes to identify optimal preventive testing strategies\colorbox{red!50}{\sout{to curb this spread . The model is}}
    \colorbox{red!50}{\sout{ microscopically.}}\colorbox{teal!50}{The model is} calibrated to high-resolution data from actual nursing homes in Austria, including the detailed networks of social contacts of their residents and information on past outbreaks.  \\
    \midrule
    2 
    & Due to its high lethality amongst the elderly, nursing homes are in the eye of the COVID-19 \colorbox{red!50}{\sout{storm}}\colorbox{teal!50}{pandemic}. Emerging new test procedures , such as antigen or RT-LAMP tests, might enable us to protect nursing home residents by means of preventive screening strategies. Here, we develop a \colorbox{red!50}{\sout{novel}}\colorbox{teal!50}{detailed} agent-based epidemiological model for the spread of SARS-CoV-2 in nursing homes to identify optimal preventive testing strategiesto curb this spread . The model is microscopically calibrated to high-resolution data from actual nursing homes in Austria, including \colorbox{red!50}{\sout{the detailed networks of social contacts of their}}
    \colorbox{red!50}{\sout{resident}}\colorbox{teal!50}{detailed social contact networks} and information on past outbreaks. 
    & \colorbox{gray!20}{\sout{Due to its high lethality amongst the elderly, n}}
    \colorbox{gray!20}{N}ursing homes are in the eye of the COVID-19 storm. Emerging new test procedures might enable us to protect nursing home residents by means of preventive screening \colorbox{red!50}{\sout{strategies }}. Here, we develop a novel agent-based epidemiological model for the spread of SARS-CoV-2 in nursing homes to identify optimal preventive testing strategies. The model is calibrated to high-resolution data from actual nursing homes in Austria, including \colorbox{red!50}{\sout{the}} detailed networks of social contacts of their residents and information on past outbreaks.  \\
    \midrule
    3 
    &  -
    & Due to its high lethality amongst the elderly, nursing homes are in the eye of the COVID-19 storm. Emerging new test procedures might enable us to protect nursing home residents by means of preventive screening. Here, we develop a\colorbox{gray!20}{\sout{ novel}}\colorbox{gray!20}{n} agent-based epidemiological model for the spread of SARS-CoV-2 in nursing homes to identify optimal preventive testing strategies. The model is calibrated to high-resolution data from actual nursing homes in Austria, including detailed networks of social contacts of \colorbox{red!50}{\sout{their}} residents and information on past outbreaks. \\
    \bottomrule
  \end{tabular}
  \caption{\label{tab:revision_example_arxiv}
  A sample snippet of iterative text revisions in ArXiv domain generated by \method, where $t$ is the revision depth and $t=0$ indicates the original input text.
  Note that \colorbox{red!50}{\sout{text}} represents user accepted deletions, \colorbox{teal!50}{text} represents user accepted insertions, and \colorbox{gray!20}{text} represents user rejected edits.
  }
\end{table*}

\paragraph{User Feedback.}
We also collected qualitative feedback about \method from the linguistic experts through a questionnaire. The first part of our questionnaire asks participants to recall their experience with the system, and evaluate various aspects of the system (in \autoref{tab:ux}). They were asked to rate how easy it was to get onboarded and use the system (\textit{convenience}), whether they were satisfied with the system (revision quality and usage experience) (\textit{satisfaction}), whether they felt it improved their productivity for text revision (\textit{productivity}), and whether they would like to use the system again (\textit{retention}) for performing revisions on their documents. 

In general, the users gave positive feedback towards the ease of use of the system. However, they were neutral on the potential productivity impact, owing to the lack of domain knowledge of the documents they were evaluating. 
This issue could be mitigated by asking users to revise their own documents of interest.
The retention and satisfaction scores were leaning slightly negative, which was explained as primarily attributed to gaps in the user interface design (eg. improperly aligned diffs, suboptimal presentation of word-level edits, etc.). 

We also asked them to provide detailed comments on their experience, and the potential impact of the system on their text revision experience. \add[VR]{Specifically, upon asking the users whether using the system to evaluate the model-suggested edits would be more time-efficient compared to actually revising the document themselves,} 
we received many useful insights that help better design better interfaces and features of our system in future work, as some users noted:

\begin{quote}
\textit{
I think it would be faster using the system, but I would still be checking the text myself in case edits were missed. 
The system made some edits where there were letters and parts of words being added/removed/replaced, which sometimes took some time to figure out. That wouldn't be the case if I were editing a document. 
}
\end{quote}

\begin{quote}
\textit{
Ultimately, I would use the system for grammar/coherence/clarity edits, and then still research (a lot) to ensure that meaning was preserved throughout the document. For topics that I was more familiar with/more general topics, using the system would probably reduce my time by a third or so. For topics that required more in-depth research for me, the time saved by using the system might be minimal.}
\end{quote}

\section{Discussion and Future Directions}
When \method generates revisions at deeper depths, we observe a decrease in the acceptance ratio by human users.  
It is crucial to create a text revision system that can learn different revision strategies at each iteration and generate high quality edits at deeper revision levels.

Editing suggestions provided by our text revision generation models could be improved. Particularly, \textsc{Fluency} edits show a huge gap between human and system revisions (45.05\% and 82.02\%).
Future work could focus on developing more powerful text revision generation models.

In our human-machine interaction, we restrict the users' role to accept or reject the model's predictions.
Even with minimal human interaction, our experiment shows comparable or even better revision quality as compared to human writers at early revision depths.
A potential future direction for human-machine collaborative text revision would be to develop advanced human-machine interaction interfaces, such as asking users to re-write the machine-revised text.\add[ZM]{

Also, a larger-scale user study could be carried out to derive more meaningful statistics (e.g. optimal number of revision depths and edit suggestions) and investigate if there is any intriguing user behavior in the iterative revision process. For example, as mentioned in the users' feedback, it would be interesting to check if users behave differently when they are asked to accept/reject edit suggestions provided for their own texts as opposed to the texts written by a third party.}

\section{Conclusion}
In this work, we develop an interactive iterative text revision system \method that is able to effectively assist users to make revisions and improve the quality of existing documents.
\method can generate higher quality revisions while minimizing the human efforts.
Users are provided with a reviewing interface to accept or reject system suggesting edits. The user-validated edits are then propagated to the next revision depth to get further improved revisions. 
Empirical results show that \method can generate iterative text revisions with acceptance rates comparable or even better than human writers at early revision depths.

\section*{Acknowledgments}
We thank all linguistic expert annotators at Grammarly for participating in the user study and providing us with valuable feedback during the process. We also thank Karin de Langis at University of Minnesota for narrating the video of our system demonstration. We would like to extend our gratitude to the anonymous reviewers for their helpful comments.

\bibliography{anthology,custom}
\bibliographystyle{acl_natbib}

\appendix
\section{\method Iterative Revision Samples}
\label{sec:appendix}

We present more iterative revision examples generated by \method in  \autoref{tab:revision_example_news} and \autoref{tab:revision_example_wiki}.

\begin{table*}[t]
  \centering
  \small
  \begin{tabular}{@{}r|p{0.5\textwidth}|p{0.4\textwidth}@{}}
    \toprule
     $t$ & \textsc{Human-Human} & \textsc{System-Human}(ours)  \\
    \midrule
    0 
    & Corporal Nathan Hornburg. A Reserve soldier serving with Canadian Forces in Afghanistanwas killed on September 24, 2007. Four others were injured in the incident which killed 24-year-old Corporal Nathan Hornburg of Calgary, Alberta. A Canadian Forces statement said Cpl. Hornburg was killed during Operation Sadiq Sarbaaz (Honest Soldier) approximately 47 kilometres west of Kandahar City in Panjwaii District. Media reports indicated he died from mortar fire at around 4 :30 p.m. local time (12:00 UTC) while he was repairing the track on a Canadian Leopard tank near a cluster of villages known as Zangabad. 
    & Corporal Nathan Hornburg. A Reserve soldier serving with Canadian Forces in Afghanistanwas killed on September 24, 2007. Four others were injured in the incident which killed 24-year-old Corporal Nathan Hornburg of Calgary, Alberta. A Canadian Forces statement said Cpl. Hornburg was killed during Operation Sadiq Sarbaaz (Honest Soldier) approximately 47 kilometres west of Kandahar City in Panjwaii District. Media reports indicated he died from mortar fire at around 4 :30 p.m. local time (12:00 UTC) while he was repairing the track on a Canadian Leopard tank near a cluster of villages known as Zangabad. \\
    \midrule
    1 
    & \colorbox{red!50}{\sout{Corporal Nathan Hornburg. A Reserve soldier serving with}}
    \colorbox{red!50}{\sout{ Canadian Forces in Afghanistanwas killed on September}}
    \colorbox{red!50}{\sout{ 24, 2007}}\colorbox{teal!50}{On MOnday, a 24-year old Calgary Reservist}
    \colorbox{teal!50}{became the 71st Canadian soldier killed in Afghanistan}. Four others were injured in the incident which killed \colorbox{red!50}{\sout{24-year-old}} Corporal Nathan Hornburg \colorbox{red!50}{\sout{of Calgary, Alberta. A Canadian Forces statement said Cpl.  }}
    \colorbox{red!50}{\sout{Hornbur was killed during}}\colorbox{teal!50}{, who was operating as part of} Operation Sadiq Sarbaaz (Honest Soldier) approximately 47 kilometres west of Kandahar City in Panjwaii District. Media reports indicated he died from mortar fire at around 4 :30 p.m. local time \colorbox{red!50}{\sout{(12:00 UTC)}} while he was repairing the track on a Canadian Leopard tank near a cluster of villages known as Zangabad. 
    & \colorbox{red!50}{\sout{Corporal Nathan Hornburg. A Reserve}}
    \colorbox{teal!50}{A Canadian} soldier serving with Canadian Forces in Afghanistanwas killed on September 24, 2007. Four others were injured in the incident which killed 24-year-old Corporal Nathan Hornburg of Calgary, Alberta. \colorbox{red!50}{\sout{A Canadian Forces statement said Cpl.}}
    \colorbox{teal!50}{Nathan} Hornburg was killed during Operation Sadiq Sarbaaz (Honest Soldier) approximately 47 kilometres west of Kandahar City in Panjwaii District. Media reports indicated he died from mortar fire at around 4 :30 p.m. local time (12:00 UTC) while he was repairing the track on a Canadian Leopard tank near a cluster of villages known as Zangabad.  \\
    \midrule
    2 
    & Corporal Nathan Hornburg. A Reserve soldier serving with Canadian Forces in Afghanistanwas killed on September 24, 2007. Four others were injured in the incident which killed 24-year-old Corporal Nathan Hornburg of Calgary, Alberta. A Canadian Forces statement said Cpl. Hornburg was killed during Operation Sadiq Sarbaaz \colorbox{red!50}{\sout{(Honest Soldier) approximately 47 kilometres west of }}
    \colorbox{red!50}{\sout{Kandahar City in Panjwaii District}}\colorbox{teal!50}{, a joint Afghan-NATO}
    \colorbox{teal!50}{ mission designed to "set the conditions for a continuous }
    \colorbox{teal!50}{security presence and the establishment of a new police }
    \colorbox{teal!50}{sub-station in the northern part of (Panjwaii)."}. Media reports indicated he died from mortar fire at around 4 :30 p.m. local time (12:00 UTC) while he was repairing the track on a Canadian Leopard tank near a cluster of villages known as Zangabad. 
    &  A Canadian soldier serving with \colorbox{teal!50}{the} Canadian Forces in Afghanistanwas killed on September 24, 2007. Four others were injured in the incident which killed 24-year-old Corporal Nathan Hornburg of Calgary, Alberta. Nathan Hornburg was killed during Operation Sadiq Sarbaaz (Honest Soldier)\colorbox{teal!50}{,} approximately 47 kilometres west of Kandahar City in Panjwaii District. Media reports indicated he died from mortar fire at around 4 :30 p.m. local time (12:00 UTC) while he was repairing the track on a Canadian Leopard tank near a cluster of villages known as Zangabad. \\
    \midrule
    3 
    & - 
    & A Canadian soldier serving with the Canadian Forces in Afghanistan was killed on September 24, 2007. Four others were injured in the incident which killed 24-year-old \colorbox{red!50}{\sout{Corporal}}\colorbox{teal!50}{Cpl.} Nathan Hornburg of Calgary, Alberta. Nathan Hornburg was killed during Operation Sadiq Sarbaaz (Honest Soldier), approximately 47 kilometres west of Kandahar City in \colorbox{teal!50}{the} Panjwaii District. Media reports indicated he died from mortar fire at around 4 :30 p.m. local time (12:00 UTC) while he was repairing the track on a Canadian Leopard tank near a cluster of villages known as Zangabad. \\
    \bottomrule
  \end{tabular}
  \caption{\label{tab:revision_example_news}
  A sample snippet of iterative text revisions in Wikinews domain generated by \method, where $t$ is the revision depth and $t=0$ indicates the original input text.
  Note that \colorbox{red!50}{\sout{text}} represents user accepted deletions, \colorbox{teal!50}{text} represents user accepted insertions, and \colorbox{gray!20}{text} represents user rejected edits.
  }
\end{table*}
\begin{table*}[t]
  \centering
  \small
  \begin{tabular}{@{}r|p{0.45\textwidth}|p{0.4\textwidth}@{}}
    \toprule
     $t$ & \textsc{Human-Human} & \textsc{System-Human}(ours)  \\
    \midrule
    0 
    & Jecon Gregory is or was a nomadic artist, whose autobiographical fragments and poems, dictated to an acquaintance, were published as the book "History of a Nation of One" (Harcourt Brace, New York, 1969, and Michael Joseph, London, 1971). Jecon apparently did not know his place, date, language or even name of birth, began his wanderings as a child in Malta; walked through many lands, barefoot, tall and thin, pulling all his possessions in a basket on wheels, sleeping on the ground, and making a living by drawing portraits.
    &  Jecon Gregory is or was a nomadic artist, whose autobiographical fragments and poems, dictated to an acquaintance, were published as the book "History of a Nation of One" (Harcourt Brace, New York, 1969, and Michael Joseph, London, 1971). Jecon apparently did not know his place, date, language or even name of birth, began his wanderings as a child in Malta; walked through many lands, barefoot, tall and thin, pulling all his possessions in a basket on wheels, sleeping on the ground, and making a living by drawing portraits. \\
    \midrule
    1 
    & Jecon Gregory is or was a nomadic artist, whose autobiographical fragments and poems, dictated to an acquaintance, were published as the book "History of a Nation of One\colorbox{gray!20}{: An Unlikely Memoir}" (Harcourt Brace, New York, 1969, and Michael Joseph, London, 1971).\colorbox{gray!20}{..} Jecon apparently did not know his place, date, language or even name of birth, began his wanderings as a child in Malta; walked through many lands, barefoot, tall and thin, pulling all his possessions in a basket on wheels, sleeping on the ground, and making a living by drawing portraits.
    & Jecon Gregory is or was a nomadic artist, whose autobiographical fragments and poems, dictated to an acquaintance, were published as the book \colorbox{gray!20}{\sout{"}}History of a Nation of One" (Harcourt Brace, New York, 1969, and Michael Joseph, London, 1971). Jecon apparently did not know his place, date, language or even name of birth, began his wanderings as a child in Malta; walked through many lands, barefoot, tall and thin, pulling all his possessions in a basket on wheels, sleeping on the ground, and \colorbox{gray!20}{\sout{making a living by}} drawing portraits. \\
    \midrule
    2 
    & -
    & Jecon Gregory is or was a nomadic artist, whose autobiographical fragments and poems, dictated to an acquaintance, were published as the book "History of a Nation of One" (Harcourt Brace, New York, 1969, and Michael Joseph, London, 1971). Jecon apparently did not know his place, date, language or even name of birth, began his wanderings as a child in Malta; walked through many lands, barefoot, tall and thin, pulling all his possessions in a basket on wheels, sleeping on the ground, and \colorbox{gray!20}{\sout{making a living by}} drawing portraits. \\
    \midrule
    3 
    & - 
    & - \\
    \bottomrule
  \end{tabular}
  \caption{\label{tab:revision_example_wiki}
  A sample snippet of iterative text revisions in Wikipedia domain generated by \method, where $t$ is the revision depth and $t=0$ indicates the original input text.
  Note that \colorbox{red!50}{\sout{text}} represents user accepted deletions, \colorbox{teal!50}{text} represents user accepted insertions, and \colorbox{gray!20}{text} represents user rejected edits.
  }
\end{table*}

\end{document}